\documentclass[times,twocolumn,final]{elsarticle}

\usepackage{medima}
\usepackage{framed,multirow}

\usepackage{amssymb}
\usepackage{latexsym}
\usepackage{amsmath}
\usepackage{algorithm}
\usepackage{algpseudocode}
\usepackage{amsthm}
\usepackage{url}
\usepackage{xcolor,colortbl}
\usepackage{booktabs} 
\usepackage[colorlinks=true]{hyperref}
\usepackage{footnote}
\usepackage{tabularx}
\makesavenoteenv{tabular}
\makesavenoteenv{table}
\usepackage{blindtext}
\usepackage{enumitem}
\usepackage{multirow}
\usepackage[labelformat=simple]{subfig}
\usepackage{threeparttable/threeparttable}
\usepackage{bm}
\usepackage[normalem]{ulem}
\usepackage{pifont}
\newcommand{\cmark}{\ding{51}}%
\newcommand{\xmark}{\ding{55}}%
\DeclareMathOperator*{\argmax}{arg\,max}

\theoremstyle{definition}
\newtheorem{definition}{Definition}[]

\definecolor{d}{RGB}{245,220,215}
\definecolor{l}{RGB}{252,240,225}

\journal{Medical Image Analysis}

\begin{document}

\verso{Tiange Xiang \textit{et~al.}}

\begin{frontmatter}

\title{Towards Bi-directional Skip Connections in Encoder-Decoder Architectures and Beyond}%

\author[1]{Tiange \snm{Xiang}}


\author[1]{Chaoyi \snm{Zhang}}

\author[1]{Xinyi \snm{Wang}}

\author[2]{Yang \snm{Song}}

\author[1]{Dongnan \snm{Liu}}

\author[3]{Heng \snm{Huang}}

\author[1]{Weidong \snm{Cai}\corref{cor1}}

\ead{tom.cai@sydney.edu.au}

\cortext[cor1]{Corresponding author.}

\address[1]{School of Computer Science, University of Sydney, Australia}
\address[2]{School of Computer Science and Engineering, University of New South Wales, Australia}
\address[3]{Department of Electrical and Computer Engineering, University of Pittsburg, USA}

\received{13 July 2021}
\accepted{10 March 2022}

\begin{abstract}
    U-Net, as an encoder-decoder architecture with forward skip connections, has achieved promising results in various medical image analysis tasks. Many recent approaches have also extended U-Net with more complex building blocks, which typically increase the number of network parameters considerably. Such complexity makes the inference stage highly inefficient for clinical applications. Towards an effective yet economic segmentation network design, in this work, we propose backward skip connections that bring decoded features back to the encoder. Our design can be jointly adopted with forward skip connections in any encoder-decoder architecture forming a recurrence structure without introducing extra parameters. With the backward skip connections, we propose a U-Net based network family, namely Bi-directional O-shape networks, which set new benchmarks on multiple public medical imaging segmentation datasets. On the other hand, with the most plain architecture (BiO-Net), network computations inevitably increase along with the pre-set recurrence time. We have thus studied the deficiency bottleneck of such recurrent design and propose a novel two-phase Neural Architecture Search (NAS) algorithm, namely BiX-NAS, to search for the best multi-scale bi-directional skip connections. The ineffective skip connections are then discarded to reduce computational costs and speed up network inference. The finally searched BiX-Net yields the least network complexity and outperforms other state-of-the-art counterparts by large margins. We evaluate our methods on both 2D and 3D segmentation tasks in a total of six datasets. Extensive ablation studies have also been conducted to provide a comprehensive analysis for our proposed methods.

\end{abstract}


\end{frontmatter}


\section{Introduction}

Accurate and efficient analysis of medical images is of great interest to the computer vision and medical communities. The diagnosis of potential disease from medical images relies on a wide range of features implied by visual clues. Such decision making process requires time-consuming efforts from physicians, slowing down the process of diagnosis. Therefore, effective and efficient computer-aided models that provide automated analysis of medical images are highly desirable.

As one of the essential medical image analysis, semantic image segmentation requires dense predictions to indicate the class of each pixel: part of a nucleus, a certain organ, or an anomaly region. Benefiting  from forward feature skips, U-Net has demonstrated its wide success in segmenting images of all kinds of modalities. However, the level-to-level forward connections are limited in feature aggregation ability and better encoder-decoder aggregation strategies are needed for more advanced feature refinement. In this work, we study the skip mechanism in encoder-decoder architectures and design effective yet efficient segmentation networks.

\subsection{Related Work}

\paragraph{Image segmentation} Computer-aided image segmentation has a long history \citep{haralick1985image} in the computer vision field. With the recent success of deep learning, neural network based methods have demonstrated their ability in fast and accurate segmentation of digital images.

As a pioneer segmentation model, U-Net \citep{ronneberger2015u} adopts an encoder-decoder structure that first encodes the visual signals provided in an image into high-dimensional features and then decodes the abstract semantics to learn the pixel-to-pixel mapping between the input image and the segmentation mask. In contrast to conventional autoencoders \citep{hinton1994autoencoders}, skip connections are added between the encoder and decoder, so that encoded features are forwarded to the decoder. Such forward skip connections enable fine-grained aggregations of features and preserve better gradient flows during network optimization. Although other advanced approaches exist in different forms \citep{chen2017deeplab, long2015fully}, U-Net often demonstrates its superior performances and serves as the most important backbone in medical image segmentation tasks.

\paragraph{U-Net variants} Recent studies have developed different building blocks to extend U-Net. For instance, V-Net \citep{milletari2016v} applies U-Net to process 3D voxel data for 3D vision tasks. W-Net \citep{xia2017w} concatenates two U-Nets head-to-tail to approach image segmentation tasks in an unsupervised style. M-Net \citep{mehta2017m} studies the impact of using multi-scale features through paired downsampling and upsampling layers. U-Net++ \citep{zhou2018unetpp}, as a direct extension of U-Net, designs nested building blocks with dense skip connections to better propagate encoded features. Apart from the modifications on network structure, attention U-Net \citep{oktay2018attention} incorporates attention mechanism into U-Net by learning self-attentions to be applied on the skipped features. However, the above U-Net variants require additional functional modules, leading to an escalation of network complexity. In our work, we improve the performance of U-Net via inserting novel backward feature skip connections, where building blocks are reused without introducing any extra network parameters.

\paragraph{Recurrent convolutional networks} Iterative refinement of features has been proved effective in many computer vision tasks \citep{han2018image,guo2019dynamic,wang2019recurrent,alom2018nuclei}. \textcolor{black}{Auto-context \citep{tu2009auto} implicitly extracts image features together with context information by learning a series of classifiers in a similar recurrent manner. However their classifiers need to be independently built in a boosting style, where our methods recurse semantic features inside a single encoder-decoder network.} \citep{guo2019dynamic} reuses ResNet residual blocks \citep{he2016deep} to fully utilize the limited parameters. With the similar recurrent strategy, \citep{wang2019recurrent} proposed R-U-Net, which connects multiple U-Net architectures head-to-tail with shared parameters to enhance segmentation performances. R2U-Net \citep{alom2018nuclei} adopts a similar approach that only recurses the last building block at each level of refinement. By contrast, our method learns recurrent bi-directional connections between encoders and decoders.
	
\paragraph{Neural architecture search} Compared to manually crafted networks, Neural Architecture Search (NAS) algorithms automatically search for the optimal architecture in a defined search space. Reinforcement Learning (RL) based methods \citep{zoph2016neural} utilize a stand-alone agent to monitor the search process under certain proxies. Evolution based methods \citep{real2017large,real2019regularized} start with a set of randomly sampled ancestor networks and then progressively evolve the population to new generations with more powerful offspring networks. Differentiable methods \citep{liu2018darts,guo2020single} optimize "architecture parameters" through back propagation by relaxing the discrete search space, and define the network topology based on the optimized architecture parameters. 

NAS searched architectures also obtained success in segmentation tasks. Auto-DeepLab \citep{liu2019auto} designed a differentiable search space to determine the best operators and topologies for each building block. Similarly, NAS-Unet \citep{weng2019unet} also adopted a gradient-based search that automatically discovers basic cell structures to construct a U-Net like architecture. As a multi-scale counterpart, the search space in \citep{yan2020ms} covers multiple encoding and decoding levels. Their proposed MS-NAS has the ability of aggregating multi-level features, which leads to better segmentation performance. However, the search process of the above methods is time consuming and computational inefficient. In this work, we design an efficient two-phase NAS method to find a subset of optimal bi-directional skip connections that yield the least network parameters and computations.

\subsection{Contributions}
 Our overall contributions include: \textbf{(1)} We propose bi-directional skip connections in encoder-decoder architectures for an iterative aggregation of encoded and decoded features; \textbf{(2)} We incorporate bi-directional skip connections into a simple U-Net architecture, namely BiO-Net, achieving better segmentation performance without using extra parameters; \textbf{(3)} The deficiency of BiO-Net is analyzed and an upgraded network, BiO-Net++, is designed with significantly less network complexity and multi-scale skip connections; \textbf{(4)} To further reduce the redundancies in BiO-Net++, we introduce an efficient two-phase BiX-NAS method to automatically search for resource-aware and optimal skip connections from the BiO-Net++. The finally searched BiX-Net model achieves on par performance to the BiO-Net but with significantly less complexity; \textbf{(5)} Our proposed bi-directional skip connections along with the novel networks were evaluated on a variety of medical image segmentation tasks including: 2D nuclei segmentation, 2D multi-organ segmentation, 3D Covid-19 infection segmentation, and two 3D segmentation tasks from the Medical Segmentation Decathlon (MSD). Our methods establish new benchmarks on these datasets; \textbf{(6)} Extensive ablation studies were conducted to fully study the usefulness of each of the proposed components.

\textcolor{black}{Compared to the preliminary versions of this work published in MICCAI 2020 \citep{xiang2020bio} and MICCAI 2021 \citep{wang2021bixnas}, our contributions made in this paper include: \textbf{(1)} Methodology is introduced with more intuitions and presented in more details through both plain texts and algorithmic presentations; \textbf{(2)} Discussions and analysis on drawbacks of our methods are provided in details; \textbf{(3)} More experiments were made for a more comprehensive evaluation on both 2D and 3D segmentation tasks; \textbf{(4)} Extensive ablative studies and analysis were conducted to better comprehend the bi-directional connections together with the progress evolutionary search.}

Our project page with source codes has been made publicly available to foster any future research \footnote{\url{https://bionets.github.io/}}.

\subsection{Symbols and Notations}

For notion simplicity, here we define several symbols and notations before presenting our methods: $\mathbf{T}$: The pre-set recurrence time. Note that $\mathbf{T}-1$ represents the total time of backward feature skipping in our bi-directional networks; \textcolor{black}{$\mathbf{N}$: The expansion multiplier for feature map dimension. For example, with a base feature dimension of 32 and $\mathbf{N}=1.25$, the expanded dimension will be 40; $\mathbf{W}$: The number of backward skip connections (start from the deepest encoding level);} $\mathbf{L}$: The greatest encoding depth; $\mathbf{P}$: The population of candidate networks of a generation; \textbf{Extraction stage}:  A sequential encoding or decoding process. There are two extraction stages (i.e., one encoder and one decoder) in U-Net; \textbf{Searching block}: Any building block in an extraction stage; $\mathbf{C}$: The ready-to-be-searched candidate skips between a pair of extraction stages; $\mathbf{E}$: The evolved skips from a set of candidate skips between a pair of extraction stages.



\begin{figure*}[t]
	\includegraphics[width=\textwidth]{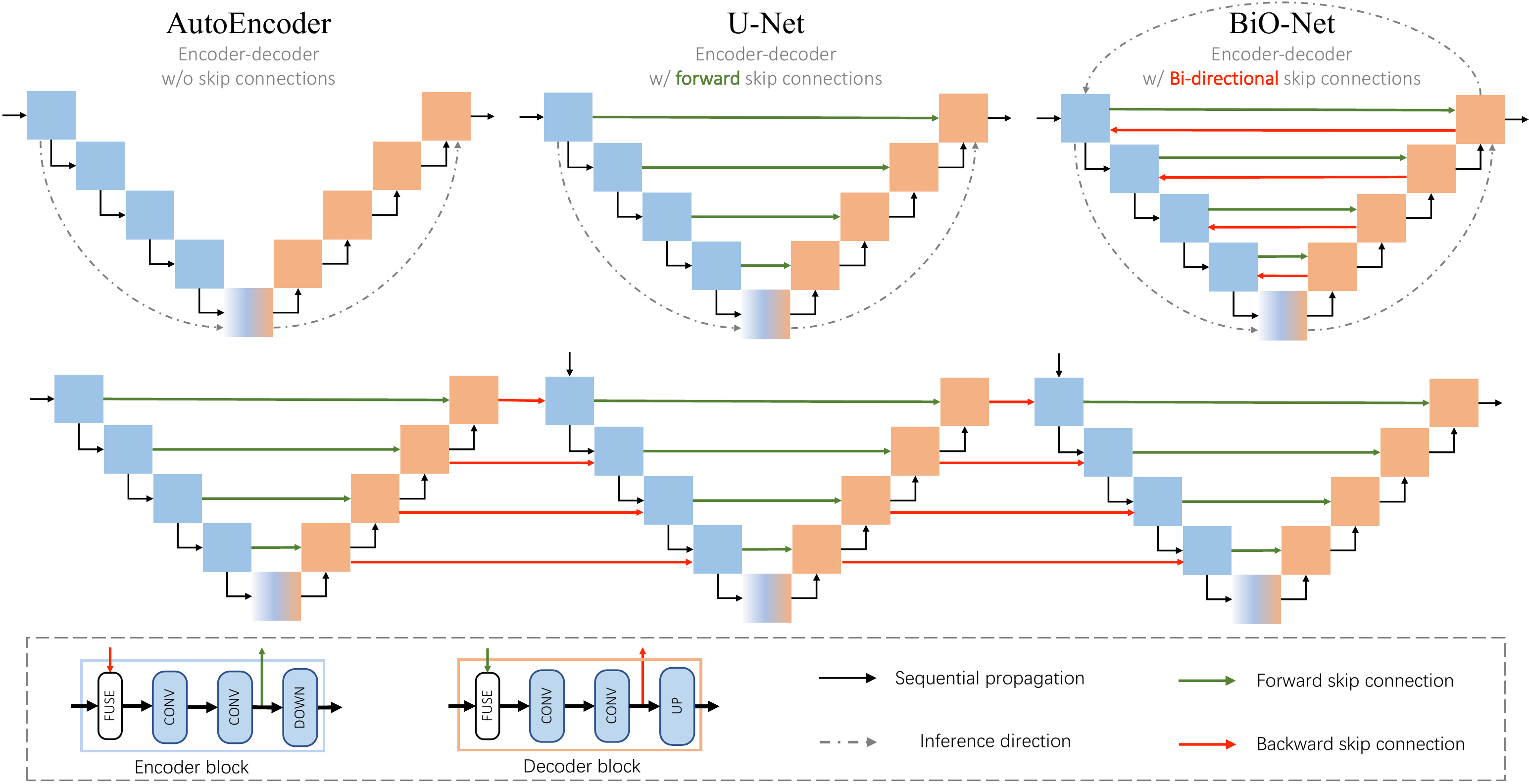}
	\caption{Top: AutoEncoder \citep{hinton1994autoencoders}, U-Net \citep{ronneberger2015u}, and our BiO-Net. With the same architecture structure, the major differences are the proposed bi-directional skip connections. Bottom: unrolled overview of our BiO-Net when \textcolor{black}{$\mathbf{T}=3$, $\mathbf{L}=4$, and $\mathbf{W}=4$}. Note that in BiO-Net, same level encoder/decoder blocks share the same weights, such that no extra parameters are required.} \label{fig1}
\end{figure*}


\section{Bi-directional Skip Connections}
Let's first consider a simple encoder-decoder architecture similar to U-Net. Skip connections are built at different levels, facilitating the information exchange between encoder and decoder.

\textcolor{black}{\paragraph{Motivation} Skip connections across different network levels have been proven effective for learning finer-grained feature representations \citep{he2016deep, ronneberger2015u}. Toward better feature extraction, networks can be designed by stacking multiple U-Nets head to tail to construct skip connections in a more structured form \citep{xia2017w}. However, with more network parameters introduced in the extra U-Nets, the introduced improvements are limited. This motivates us to explore skip connections through a recurrent manner to not expand networks further but reuse parameters instead. In this way, we argue that semantic features can be skipped bidirectionally between both encoder and decoder. Our design aims at not only promoting feature aggregations better but also fully utilizing only the parameters in a single encoder-decoder network.} 

\paragraph{Forward Skip Connections} Forward skip connections carry forward the encoded low-level features to the decoder. There are two incoming feature streams to each decoder block: the sequential feature stream from a lower level decoder block $\hat{\textbf{x}}_{in}$ and the forward skipped feature stream $\textbf{f}_{enc}$ from the corresponding encoder block at the same level. The decoder block then fuses the two streams through a $\texttt{FUSE}$ operation, and the features are subsequently propagated through the decoding convolutions $\texttt{DEC}$ to generate semantic features $\textbf{f}_{dec}$. This process can be defined as:
\begin{equation} \label{eq1}
\textbf{f}_{dec} = \texttt{DEC}(\texttt{FUSE}(\textbf{f}_{enc},\ \hat{\textbf{x}}_{in})).
\end{equation}  
\noindent
\paragraph{Backward Skip Connections} Paired to forward skip connections, we define the backward skip connections that pass the decoded high-level semantic features $\textbf{f}_{dec}$ back to the encoder. An encoder block now combines $\textbf{f}_{dec}$ with its original sequential input $\textbf{x}_{in}$ from an upper level encoder block to create additional aggregations between low-level and high-level features. Similar to \ref{eq1}, our encoding process can be formulated with its encoding convolutions $\texttt{ENC}$ as:
\begin{equation} \label{eq2}
\textbf{f}_{enc} = \texttt{ENC}(\texttt{FUSE}(\textbf{f}_{dec}, \ 
\textbf{x}_{in})).
\end{equation}

\subsection{Recurrent Inference}
Unlike forward skips, our proposed backward skip connections cannot work directly in the classic one-pass encoder-decoder architecture, since encoder layers will only be forwarded once. To enable backward feature skipping, we design an "O"-shape recurrent inference routine for bi-directional networks (Sec. \ref{sec3}) that propagate features iteratively between encoder and decoder through the bi-directional skip connections. Noticeably, such recurrent inference reuses existing network weights, and no extra parameters are introduced during the inference. Finally, the building block outputs using our bi-directional skip connections can be formulated as follows, at the iteration $i$:
\begin{equation}
\begin{split}
\textbf{x}_{out}^i &= \texttt{DOWN}(\texttt{ENC}(\texttt{FUSE}(\texttt{DEC}(\texttt{FUSE}(\textbf{f}_{enc}^{i-1}, \hat{\textbf{x}}_{in}^{i-1})), \textbf{x}_{in}^i))),\\
\hat{\textbf{x}}_{out}^i &= \texttt{UP}(\texttt{DEC}(\texttt{FUSE}(\texttt{ENC}(\texttt{FUSE}(\textbf{f}_{dec}^i, \ 
\textbf{x}_{in}^i)),\hat{\textbf{x}}_{in}^i))),
\end{split}
\end{equation}

\noindent
where \texttt{DOWN} represents the downsampling process, and \texttt{UP} represents the upsampling process.

\section{Recurrent Bi-directional Networks} \label{sec3}

\begin{figure*}[t]
	\includegraphics[width=\textwidth]{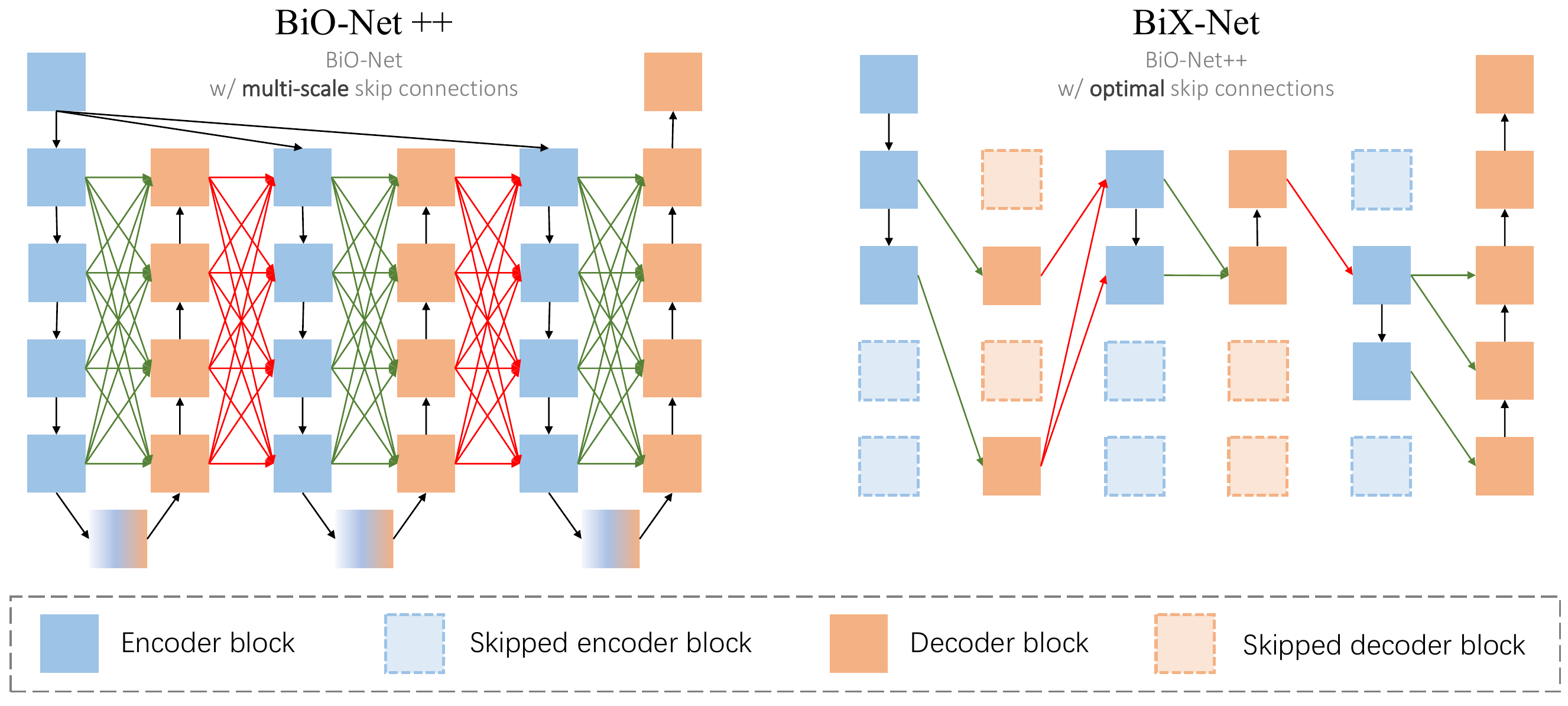}
	\caption{Unrolled overview of our BiO-Net++ with $\mathbf{T}=3$ and the finally searched BiX-Net. Building blocks are skipped if there is no connected path linking them to the post-processing block. Compared to the densely connected BiO-Net++, the sparser BiX-Net is more memory economic and computational efficient.} \label{nas_nets}
\end{figure*}

\begin{figure}[t]
	\includegraphics[width=\linewidth]{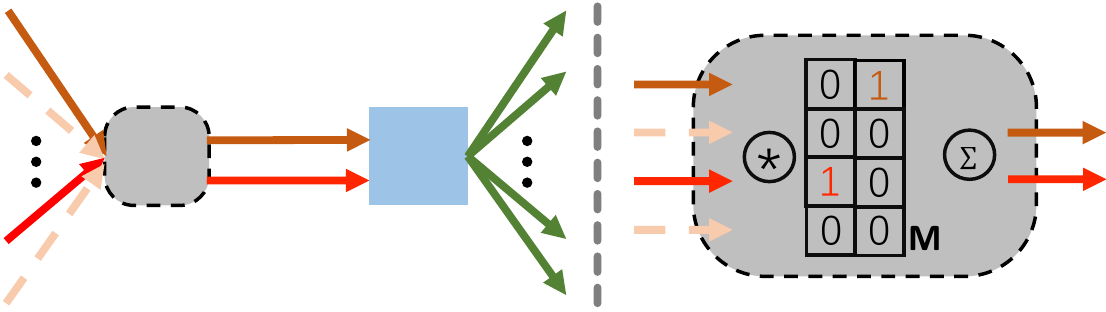}
	\caption{The proposed selection matrix strategy with an exemplary encoder block. Two skips are selected from four incoming backward skips by the learnable one-hot matrix $\mathbf{M}$.} \label{selection}
\end{figure}

\subsection{BiO-Net: A Bi-directional U-Net}

Our bi-directional skip connections are believed to be universally applicable. In this study, we insert them into a basic encoder-decoder architecture, namely BiO-Net, for method developments and evaluations. BiO-Net involves only the plain convolutional layers, batch normalization \citep{ioffe2015batch} layers, and ReLU \citep{nair2010rectified} activation layers. Note that no batch normalization layer is reused through network recursion. To align with U-Net, our BiO-Net adopts max pooling, convolution transpose and concatenation as $\texttt{DOWN}$, $\texttt{UP}$ and $\texttt{FUSE}$, respectively. An overview of our BiO-Net is shown in Fig. \ref{fig1}.

Giving an input image, we first apply three convolution-normalization-activation combinations to extract low-level features. There is no skip connection attached to such pre-processing blocks and the parameters will not be reused. The initially extracted features are then sent to a cascade of encoder blocks that are recursed through the bi-directional skip connections. Note that no features are backward skipped at the first recurrence. To make the feature map size consistent along different iterations, we duplicate the encoded features at the first encoding iteration.  

After the encoding stage, a bridge block with additional convolutions is employed to transfer the encoded features. Subsequently, a series of decoder blocks consume the features and recover encoded details using convolution transpose. During the decoding stage, our backward skip connections preserve retrieved features by concatenating them with the ones skipped from the same level encoder block. The recursion begins at the end of the last decoder block. After recursing for $\mathbf{T}$ iterations, the decoded output will be fed into the post-processing block constructed similarly to the pre-processing block. The post-processing blocks will not be involved in the recurrence. 

Optimizing a recurrent network can be tricky, since multiple computation graphs are possibly involved in one network structure. In classic Recurrent Neural Networks (RNN) \citep{hochreiter1997long}, BackPropagation Through Time (BPTT) strategy is used to calculate gradients based on outputs at different time stamps. However, in our bidirectional networks, the blocks at the finest-resolution are not recursable and only one final segmentation mask is predicted during the entire inference regardless of recurrence time. To this end, the gradient computation graph remains consistent to an unrolled version of BiO-Net (Fig. \ref{fig1}, bottom). During backpropagation, the gradients are accumulated at each learnable layer and are subsequently used to update the weights for only once.

\subsection{Analysis of Computational Burdens in BiO-Net} \label{analysis}

For a fair validation of the proposed bi-directional skip connections, our BiO-Net structure is constructed identically to U-Net with the same upsampling, fusion and channel expansion strategies. However, we find that the identical setting puts unexpected computation burdens on our BiO-Net and there exists efficient alternatives to reduce the overall complexity.

When using concatenation as the fusion strategy in BiO-Net, the encoder feature channels have to be doubled for carrying additional backward skipped features. We optimize such discord by replacing channel-wise concatenation to element-wise average pooling as the fusion strategy. Moreover, instead of using convolution transpose for upsampling with extra learning parameters, we bilinearly resize the coarse-scale feature maps to be aligned with the finer ones. As a result, the above simple optimizations reduce the complexity of BiO-Net with a 34.86 times reduction on the network parameters. 

The optimization strategies are adopted in our other networks including BiO-Net++ and BiX-Net, which will be presented in detail shortly.

\subsection{BiO-Net++: A Multi-scale Upgrade of BiO-Net}

Multi-scale information provides better guidance for the analysis of various size objects \citep{yan2020ms}. To fuse multi-scale features in BiO-Net, precedent encoded/decoded features at all levels are densely connected to every decedent decoding/encoding level through the proposed bi-directional skip connections. We average over the multi-scale features and align inconsistent spatial dimensions via simple bilinear resizing. The suggested multi-scale BiO-Net++ is outlined in Fig. \ref{nas_nets}, left.

\section{Search for Efficient Multi-scale BiO-Net}

\subsection{BiX-NAS: Two-phase Search for Efficient BiO-Net++} \label{exnas}

Although BiO-Net++ promotes multi-scale feature fusion, empirically, we found that the \textit{dense} skip connections bring marginal improvements in terms of the overall performance (Table \ref{tab:ablation4}). A natural question hence arises about whether every level-to-level skip in BiO-Net++ carries indicative information and if there exists a subset of \textit{sparse} skip connections that not only benefit from multi-scale feature fusions but yield the least complexity.

To this end, we present BiX-NAS, a two-phase NAS algorithm, to automatically find optimal and sparsely connected skip connections in the BiO-Net++. In Phase1, we follow a differentiable NAS strategy that aims at narrowing down the huge search space swiftly. In Phase2, we design an evolution-based NAS to progressively discover the best skips with the optimal segmentation performance and computation efficiency. In this study, we searched the final architecture on one dataset only and the same architecture is then transferred to all other tasks.

\begin{figure*}[t]
	\includegraphics[width=\textwidth]{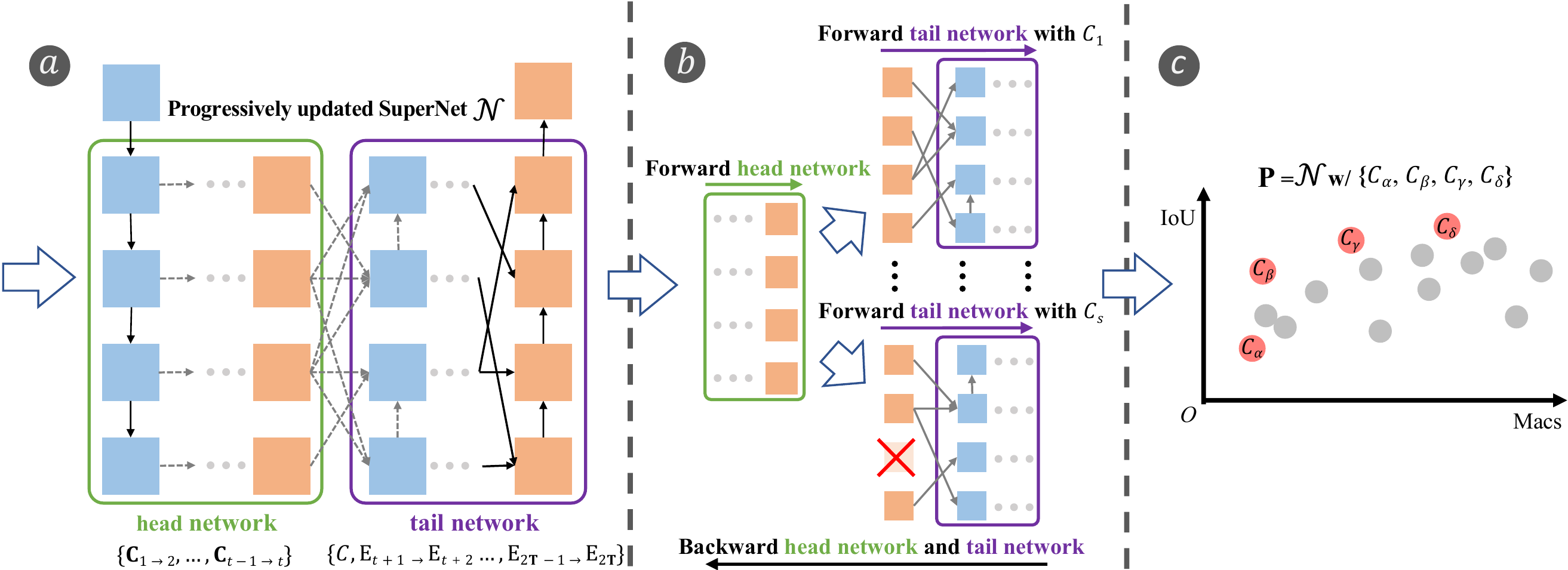}
	\caption{Our progressive evolutionary search (Phase2) workflow. \textbf{(a)} Phase1 searched SuperNet $\mathcal{N}$ can be divided into \textit{head network} and \textit{tail network}. SuperNet $\mathcal{N}$ is sustainable to be updated with the newly evolved skips. \textbf{(b)} Proposed forward and backward schemes for head and tail networks. Only one backward pass is required to update all architecture instances. \textbf{(c)} Only the searched skips at the Pareto front of a population $\mathbf{P}$ are retained and used to update $\mathcal{N}$.} \label{search}
\end{figure*}

\paragraph{Phase1: Narrowing down search space via selection matrix} 

To identify a sparse set of skip connections, the dense ones between every pair of extraction stages are evaluated and sifted. Suppose there are $N$ incoming feature streams in a desired searching block, we anticipate that only $k \in [1, N-2]$ candidate(s) of them could be accepted, which results in a search space of $\approx\sum_{k=1}^{(N-2)}\binom{N}{k}^{\mathbf{L}(2\mathbf{T}-1)}$ in the SuperNet BiO-Net++ with $\mathbf{L}$ levels and $\mathbf{T}$ iterations. When $N=5, \mathbf{L}=4$ and $\mathbf{T}=3$ (a common setup), the search space expands to $5^{40}$, escalating the searching difficulty for one optimal instance.

To alleviate such difficulty, we determine $k$ \textit{candidate skips} from $N$ incoming skips in each searching block by reducing the easy-to-spot ineffective ones. Intuitively, one-to-one relaxation parameters $\alpha$ \citep{liu2018darts} for each skip connection $x$ could be registered and optimized along with BiO-Net++. The skip with the highest $\alpha$ is then picked as the output of 
$\mathbf{\Phi}(\cdot)$, such that $\mathbf{\Phi}(\cdot) =x_{\argmax\boldsymbol{\alpha}}$, where $\mathbf{x}=\{x_1,\cdots,x_N\}$ and $\boldsymbol{\alpha}=\{\alpha_1, \cdots, \alpha_N\}$ denote the full set of incoming skips, and their corresponding relaxation scores, respectively. 

The above formulation outputs a fixed number of skips for every level. However, different levels may fuse different number of skips and the above intuitive formulation cannot suffice. Towards a more flexible skip selection, we construct a learnable \textit{selection matrix} $\mathbf{M}\in \mathbb{R}^{N \times (N-2)}$ that models the mappings between the $N$ incoming skips and $k$ anticipated candidates, and formulate $\mathbf{\Phi}(\cdot)$ as a fully differentiable equation below:

\begin{equation} \label{selectionmatrix}
    \mathbf{\Phi}(\mathbf{x},\mathbf{M})=Matmul(\mathbf{x}, \texttt{GumbelSoftmax}(\mathbf{M})),
\end{equation} 
\noindent
where the \texttt{GumbelSoftmax} trick \citep{jang2016categorical} forces each of the $(N-2)$ columns of $\mathbf{M}$ to be an one-hot vector that votes for one of the $N$ incoming skips. Our formulation generates $(N-2)$ selected skips with repetition allowed, achieving a flexible selection of $[1, N-2]$ different candidate skips (Fig. \ref{selection}). The \textit{unique} candidate skips are then averaged out to be fed into subsequent blocks. Moreover, differing from \citep{liu2018darts,liu2019auto}, our $\mathbf{\Phi}(\cdot)$ formulation unifies the forward propagation behaviour during both searching and network inference stages.

\paragraph{Phase2: Progressive evolutionary search} After squeezing the initial search space, evolving randomly sampled candidate network instances becomes practical. To further reduce skip redundancies and identify the best network instance, we perform an additional evolutionary search to find the optimal skip set at all levels across all iterations. Specifically, we search the candidate skips $\mathbf{C}$ for all levels at the same time between a certain pair of adjacent extraction stages, and then progressively move to the next pair once the current search is concluded. As the connectivity of adjacent extraction stages depends on the connectivity of precedent ones, we initiate the search from the last extraction stage pair and progressively move to the first pair. 

The conventional evolutionary NAS algorithms optimize the SuperNet with each sub-network in a population $\mathbf{P}$ individually \citep{real2019regularized,real2017large}, and then update $\mathbf{P}$ when all individual training finish. When searching for skip connections rather than operators, there are two major flaws of such strategy: first, optimizing SuperNet with sampled skip sets individually may result in unfair outcomes; second, the searching process could be empirically slow. Assuming the forward and backward of each extraction stage takes $\mathbf{I_{F}}$ and $\mathbf{I_{B}}$ time, training all $|\mathbf{P}|$ instances individually per step takes $2\mathbf{T|P|(I_{F}+I_{B})}$ in total.

\paragraph{Improving searching fairness and efficiency} \label{analysis}
To overcome the first flaw above, we define the concept of \textit{skip fairness} and claim that all skip search algorithms need to meet such principle.

\begin{algorithm}[t] 
		\caption{Progressive evolutionary search}
		\label{alg1}
		\begin{algorithmic}[1]
			\Require Iteration $\mathbf{T}$, sampling number $\mathbf{s}$, randomly initialized SuperNet weights $\mathcal{W}$, Phase1 searched \textbf{\underline{c}}andidate skips $\{\mathbf{C}_{1\to2}, \cdots, \mathbf{C}_{2\mathbf{T}-1\to2\mathbf{T}}\}$, criterion $\mathcal{L}$. 
			\Ensure BiX-Net with \textbf{\underline{e}}volved skips $\{{\mathbf{E}}_{1\to2}, \cdots, {\mathbf{E}}_{2\mathbf{T}-1\to2\mathbf{T}}\}$
			\For{$t=2\mathbf{T}-1,\cdots,1$}
			\For{$i=1,\cdots,\mathbf{s}$}
			\For{each searching block $b$}
			\State Randomly sample $n$ skips from $\mathbf{C}_{t\to t+1}^b$: $C^b_i, 1\leq n \leq|\mathbf{C}_{t\to t+1}^b|$.
			\EndFor
			\EndFor
			\For{data batch X, target Y}
			\State Forward the head network with candidate skips $\mathbf{C}_{1\to2}, \cdots, \mathbf{C}_{t-2\to t-1}$.
			\For{$i=1,\cdots,\mathbf{s}$}
			\State Forward each tail network with sampled skips $C_i$ and previously evolved skips ${\mathbf{E}}_{t+1\to t+2},\cdots,{\mathbf{E}}_{2\mathbf{T}-1\to 2\mathbf{T}}$.
			\State Calculate loss $l_i=\mathcal{L}$(X, Y)
			\EndFor
			\State Optimize $\mathcal{W}$ with the average loss  $\frac{1}{\mathbf{s}}\sum_{i=1}^{\mathbf{s}}l_i$.
			\EndFor
			\State Get Pareto front from $\{C_1,\cdots,C_\mathbf{s}\}$ and determine ${\mathbf{E}}_{t\to t+1}$.
			\EndFor
		\end{algorithmic}
		
\end{algorithm}

\begin{table*}[t]
		\caption{Details of the datasets used in our experiments. For CHAOS, number of data is reported for both 3D volumes and 2D slices.}\label{dataset}
		\centering
		\setlength\tabcolsep{0.6em}
		\begin{tabular}{{c|c |c |c |c | c | c  }}
			\toprule
			\multicolumn{1}{c}{}&\multicolumn{1}{c}{MoNuSeg} & \multicolumn{1}{c}{TNBC} &\multicolumn{1}{c}{CHAOS} & \multicolumn{1}{c}{Covid-19} & \multicolumn{1}{c}{Heart} & \multicolumn{1}{c}{Hippocampus} \\
			\hline
			\hline
			\#training data &  30 &  - & 120(1594) & 20 &  20 & 260 \\
			\#testing data  &  14 & 50 & - & - & - & - \\
			Evaluation$^1$ & train/val & val & CV & CV & CV & CV  \\ 
			Input size & $512\times512\times3$ & $512\times512\times3$& $256\times256\times1$ & $160\times160\times80$ & $192\times128\times80$ & $56\times40\times40$ \\
			Modality &microscopy&microscopy&MRI&CT&MRI&MRI\\
			\#classes & 2 & 2 & 5 & 3 & 2 & 3\\
			Batch size & 2 & 2 & 16 & 2 & 2 & 9 \\
			\bottomrule
		\end{tabular}
		\footnotesize
		\begin{tablenotes}
		\setlength{\itemindent}{0.in}
		    \item $^{1}$ Evaluation strategies. CV denotes 5-fold cross validation on all training data.
		\end{tablenotes}
\end{table*}


\begin{definition}
\textbf{Skip Fairness.} Let $\mathbf{F} = \{\mathbf{f}^1,\cdots,\mathbf{f}^N\}$ be the incoming skipped features to any searching blocks in each evolving architecture $\mathcal{A}^i$ within a population $\mathbf{P}$. The skip fairness requires $\mathbf{f}^1_{\mathcal{A}^1}\equiv\cdots\equiv\mathbf{f}^1_{\mathcal{A}^{|\mathbf{P}|}},\cdots,\mathbf{f}^N_{\mathcal{A}^1}\equiv\cdots\equiv\mathbf{f}^N_{\mathcal{A}^{|\mathbf{P}|}}$ $\forall \mathbf{f}^i \in \mathbf{F}, \forall \mathcal{A}^i \in \mathbf{P}$.
\end{definition}

The above principle yields that, when searching between the same extraction stage pair, any corresponding level-to-level skips across different sampled architectures are required to carry identical features. Otherwise, the inconsistent incoming features would impact the search decision on the skipping topology, hence causing unexpected search unfairness. 
Gradient-based search algorithms (e.g., Phase1 search algorithm) meet this principle by its definition, as the same forwarded features are distributed to all candidate skips equally. 
However, the aforementioned conventional strategy violates such principle due to the inconsistent incoming features produced by the individually trained architectures.

Our proposed Phase2 search algorithm meets the skip fairness by synchronizing partial forwarded features in all sampled candidate networks. Specifically, suppose we are searching skips between the $t^{th}$ and $t+1^{th}$ extraction stages $(t\in[1, 2\mathbf{T}-1])$: network topology from the $1^{st}$ to $t-1^{th}$ stages is fixed and the forward process between such stages can be shared. We denote such stages as \textit{head network}. On the contrary, network topology from the $t^{th}$ to $2\mathbf{T}^{th}$ stages varies as the changes of different sampled skips. We then denote such unfixed stages as \textit{tail networks}, which share the same SuperNet weights but with distinct topologies. The forwarded features of head network are fed to all candidate tail networks individually, as shown in Fig. \ref{search}. We average the losses of all tail networks, and backward the gradients through the SuperNet weights only once. Besides, our Phase2 searching process is empirically efficient and overcomes the second flaw above, as one-step training only requires $\mathbf{I_B}+\sum_{t=1}^{2\mathbf{T}-1}(t\mathbf{I_F}+(2\mathbf{T}-t)\mathbf{I_F\cdot|P|})$.

After the search between an extraction stage pair completes, we follow a multi-objective selection criterion that retains the architectures on the Pareto front \citep{yang2020cars} based on both validation accuracy (IoU) and computational complexity (MACs). The proposed progressive evolutionary search details are presented in Algorithm \ref{alg1} and our finally searched BiX-Net is shown in Fig. \ref{nas_nets}, right.

\section{Experimental Setup}
\subsection{Datasets}
\textcolor{black}{We evaluated our methods on 4 tasks including 6 public 2D and 3D datasets.} Each task requires distinct segmentation targets and represents a different imaging modality. Details of the datasets are presented in Table \ref{dataset}.

\subsubsection{MoNuSeg and TNBC} 

The MoNuSeg dataset \citep{kumar2017dataset} consists of tissue image tiles at $40\times$ magnification level from the TCGA \citep{tomczak2015cancer} database. The original tissues were sampled from multiple patients with tumors in a variety of organs from different clinics. There are a total of 44 images each of $1000\times1000$ pixels, and they are divided into a training set of 30 images and a test set of 14 images. Following a common protocol \citep{graham2019hover}, we extracted $512\times 512$ patches at 4 corners of each image, which enlarges the dataset by 4 times. Neither numeric normalization nor stain normalization was performed as pre-processing on the dataset.

The TNBC dataset \citep{naylor2018segmentation} contains 50 sampled tiles from breast tissues with a size of $512\times512$ pixels. There is no specific training-testing split available for TNBC. Compared to the data in MoNuSeg, TNBC tiles were extracted from considerably independent sources and were processed with different staining and color adjustment strategies. Therefore, we used TNBC as an extra validation set to evaluate the generalization ability of our proposed method on nuclei segmentation task by training on the MoNuSeg dataset only. 

\subsubsection{CHAOS}

The CHAOS (Combined Healthy Abdominal Organ Segmentation) dataset \citep{kavur2021chaos} consists CT and MRI volumes of different organs including liver, left kidney, right kidney and spleen. Similar to \citep{yan2020ms}, we use the training MRI image slices in this work to evaluate our methods on 2D multi-class organ segmentation task. 

There are two types of MRI sequences in the dataset with each consists of 120 DICOM volumes: T1-DUAL (40 data for both in phase and out phase) and T2-SPIR (40 data). Each volume has being performed to scan abdomen under different radio frequency pulse and gradient combinations. The datasets are acquired by a 1.5T Philips MRI, which produces the 12 bit DICOM images with $256\times256$ resolution. The ISDs vary between 5.5-9 mm (average 7.84 mm), x-y spacing is between 1.36 - 1.89 mm (average 1.61 mm) and the number of slices is between 26 and 50 (average 36). In total, there is 1594 slices (532 slice per sequence) in the dataset. Before being processed by the networks, we performed additional pre-processing techniques on the raw sequences by min-max normalization and histogram auto-contrast that extend values to span over [0, 1].

\begin{table*}[t]
		\caption{Comparison on MoNuSeg testing set and TNBC. \textbf{AS} denotes if the network is automatically searched. \textbf{RC} denotes if the network is recurrent. Highlighted cells represent the results significantly lower than BOTH BiO-Net and BiX-Net based on the statistical test (p-value $<$ 0.05). For each metric, the best result is in bold and the second best result is in underline.
} \label{comp:nuclei}
		\centering
		\setlength\tabcolsep{0.9em}
		\begin{tabular}{{l | c c |c c |c c |c | c}}
			\toprule
			 \multicolumn{3}{c}{}&\multicolumn{2}{c}{MoNuSeg} & \multicolumn{2}{c}{TNBC} \\
            \hline
			\textbf{Methods} & \textbf{AS} & \textbf{RC} &  \textbf{IoU (\%)} & \textbf{DICE (\%)}  & \textbf{IoU (\%)} & \textbf{DICE (\%)}  &\textbf{\#Params}& \textbf{MACs}$^1$\\
			\hline
			\hline
			U-Net     & \xmark & \xmark & \cellcolor{d}68.2$\pm$0.3 & \cellcolor{d}80.7$\pm$0.3 & \cellcolor{d}46.7$\pm$0.6 & \cellcolor{d}62.3$\pm$0.6 & 8.64 M & 65.83 G  \\
			U-Net++   & \xmark & \xmark & \cellcolor{d}69.4$\pm$0.3 & \cellcolor{d}81.5$\pm$0.4 & \cellcolor{d}53.9$\pm$0.4 & \cellcolor{d}67.2$\pm$0.5 & 9.16 M & 138.60 G \\
			Att U-Net & \xmark & \xmark & \cellcolor{d}68.3$\pm$0.2 & \cellcolor{d}81.1$\pm$0.2 & \cellcolor{d}56.4$\pm$0.5 & \cellcolor{d}69.9$\pm$0.6 & 8.73 M& 66.97 G\\
			\hline
			R-UNet  ($\mathbf{T}=2$)  & \xmark&\cmark& \cellcolor{d}68.5$\pm$0.2 & \cellcolor{d}80.8$\pm$0.2 & \cellcolor{d}53.7$\pm$0.4 & \cellcolor{d}66.0$\pm$0.6& 4.50 M & 103.24 G \\
			R-UNet  ($\mathbf{T}=3$)  & \xmark&\cmark& \cellcolor{d}68.8$\pm$0.3 & \cellcolor{d}81.1$\pm$0.2 & \cellcolor{d}55.5$\pm$0.5 & \cellcolor{d}68.3$\pm$0.5 & 4.50 M & 154.86 G \\
			R2U-Net ($\mathbf{T}=2$)  & \xmark&\cmark& \cellcolor{d}68.8$\pm$0.4 & \cellcolor{d}80.9$\pm$0.3 & \cellcolor{d}56.2$\pm$0.6 & \cellcolor{d}68.9$\pm$0.7 & 9.78 M & 152.89 G\\
			R2U-Net ($\mathbf{T}=3$)  & \xmark&\cmark& \cellcolor{d}69.1$\pm$0.3 & \cellcolor{d}81.2$\pm$0.3 & \cellcolor{d}60.1$\pm$0.5 & \cellcolor{d}71.3$\pm$0.6 & 9.78 M & 197.16 G \\
			\hline
			NAS-UNet    & \cmark& \xmark& \cellcolor{d}68.4$\pm$0.3& \cellcolor{d}80.7$\pm$0.3 & \cellcolor{d}54.5$\pm$0.6 & \cellcolor{d}69.6$\pm$0.5& \underline{2.42 M} &67.31 G \\
			AutoDeepLab & \cmark& \xmark& \cellcolor{d}68.5$\pm$0.2& \cellcolor{d}81.0$\pm$0.3 & \cellcolor{d}57.2$\pm$0.5 & \cellcolor{d}70.8$\pm$0.5 &27.13 M & \underline{60.33 G} \\
			MS-NAS      & \cmark& \xmark& \cellcolor{d}68.8$\pm$0.4& \cellcolor{d}80.9$\pm$0.3 & \cellcolor{d}58.8$\pm$0.6 & \cellcolor{d}71.1$\pm$0.5&14.08 M&72.71 G\\
			\hline
			BiO-Net ($\mathbf{T}=3$) & \xmark&\cmark& \textbf{69.9$\pm$0.2} &  \underline{82.0$\pm$0.2} &  \underline{62.2$\pm$0.4} &  \underline{75.8$\pm$0.5} &  14.99 M &  115.67 G\\
			BiX-Net & \cmark&\cmark& \underline{69.9$\pm$0.3} & \textbf{82.2$\pm$0.2} & \textbf{68.0$\pm$0.4} & \textbf{80.8$\pm$0.3} & \textbf{0.38 M}& \textbf{28.00 G} \\
			\bottomrule
		\end{tabular}
		\footnotesize
		\begin{tablenotes}
		\setlength{\itemindent}{0.in}
		    \item $^{1}$ MACs are calculated based on the input size of $512\times512\times3$.
		\end{tablenotes}
	\end{table*}
	
	\begin{table*}[t]
		\caption{Quantitative comparison on CHAOS MRI. Darker highlighted cells represent the results are significantly lower than BOTH BiO-Net and BiX-Net (p-value $<$ 0.05). Lighter highlighted cells represent the results are significantly lower than BiO-Net (p-value $<$ 0.05). For each metric, the best result is in bold and the second best result is in underline.}\label{comp:chaos}
		\centering
		\setlength\tabcolsep{0.6em}
		\begin{tabular}{{l|c c|c c|c c|c c}}
			\toprule
			\multicolumn{1}{c}{} & \multicolumn{2}{c}{Liver} & \multicolumn{2}{c}{Left Kidney}& 	\multicolumn{2}{c}{Right Kidney}&\multicolumn{2}{c}{Spleen}\\
			\hline
			Methods &  mIoU (\%) & DICE (\%) & mIoU (\%) & DICE (\%) & mIoU (\%)&DICE (\%) & mIoU (\%) & DICE (\%) \\
			\hline
			\hline
			U-Net       & \cellcolor{d}78.1$\pm$2.0 & \cellcolor{d}86.8$\pm$1.8 & \cellcolor{d}61.3$\pm$1.1 & \cellcolor{d}73.8$\pm$1.2 & \cellcolor{d}63.5$\pm$1.1 & \cellcolor{d}76.2$\pm$1.1 & \cellcolor{d}62.2$\pm$2.1 & \cellcolor{d}74.4$\pm$2.3\\
			U-Net++     & \cellcolor{d}78.8$\pm$1.7 & \cellcolor{d}87.5$\pm$1.6 & \cellcolor{d}65.1$\pm$1.3 & \cellcolor{d}74.7$\pm$1.4 & \cellcolor{d}66.0$\pm$1.3 & \cellcolor{d}77.2$\pm$1.7 & \cellcolor{d}64.4$\pm$1.6 & \cellcolor{l}75.6$\pm$1.5\\
			\hline
			R-UNet      & \cellcolor{d}78.2$\pm$2.1 & \cellcolor{d}86.7$\pm$1.9 & \cellcolor{d}63.6$\pm$1.5 & \cellcolor{d}75.0$\pm$1.6 & \cellcolor{d}64.4$\pm$1.3 & \cellcolor{d}75.8$\pm$1.7 & \cellcolor{d}63.4$\pm$2.0 & \cellcolor{l}75.3$\pm$1.4\\
			R2U-Net     & \cellcolor{d}77.9$\pm$1.9 & \cellcolor{d}86.4$\pm$1.8 & \cellcolor{d}63.7$\pm$1.7 & \cellcolor{d}75.1$\pm$1.6 & \cellcolor{d}64.6$\pm$1.9 & \cellcolor{d}76.1$\pm$1.9 & \cellcolor{d}63.2$\pm$1.9 & \cellcolor{d}75.0$\pm$1.5\\
			\hline
			NAS-UNet    & \cellcolor{d}79.1$\pm$1.8 & \cellcolor{d}87.2$\pm$1.8 & \cellcolor{d}65.5$\pm$1.5 & \cellcolor{d}75.0$\pm$1.3 & \cellcolor{d}66.2$\pm$1.2 & \cellcolor{d}77.7$\pm$1.0 & \cellcolor{d}64.1$\pm$1.3 & \cellcolor{l}75.8$\pm$1.6 \\
			AutoDeepLab & \cellcolor{d}79.8$\pm$1.9 & \cellcolor{d}88.1$\pm$1.8 & \cellcolor{d}66.7$\pm$1.6 & \cellcolor{d}75.0$\pm$1.7 & \cellcolor{d}61.9$\pm$0.9 & \cellcolor{d}75.7$\pm$1.1 & \cellcolor{d}63.9$\pm$1.2 & \cellcolor{l}75.5$\pm$1.4\\
			MS-NAS      & \cellcolor{d}72.6$\pm$2.3 & \cellcolor{d}82.6$\pm$2.1 & \cellcolor{l}71.0$\pm$1.3 & \cellcolor{l}81.9$\pm$1.3 & \cellcolor{l}70.1$\pm$1.9 & \cellcolor{l}81.1$\pm$1.8 & \cellcolor{d}62.5$\pm$2.1 & \cellcolor{d}74.0$\pm$2.3\\
			\hline
			BiO-Net     &  \textbf{85.8$\pm$2.0}& \textbf{91.7$\pm$1.8} & \textbf{75.7$\pm$1.1}& \textbf{85.1$\pm$1.2}& \textbf{78.2$\pm$1.0}& \textbf{87.2$\pm$1.1}& \textbf{73.2$\pm$2.3}&\textbf{82.8$\pm$2.3}\\
			BiX-Net     & \underline{82.6$\pm$1.5}&\underline{89.8$\pm$1.5}&\underline{71.0$\pm$1.0}&\underline{82.1$\pm$1.1}&\underline{71.9$\pm$0.8}&\underline{82.7$\pm$1.0}&\underline{66.0$\pm$1.7}&\underline{76.5$\pm$2.0}\\
			\bottomrule
		\end{tabular}
	\end{table*}

\subsubsection{Covid-19}

We used the most recent Covid-19 chest CT datasets \citep{ma2020towards} to validate our methods on 3D segmentation tasks. Each of the 20 3D volumes in the dataset has been annotated into 4 classes: background, left lung, right lung, and Covid-19 infection. With focus on the segmentation of Covid-19 infection regions, following \citep{muller2020automated} we combine both lung classes, resulting in a 3-class segmentation task. The CT volumes have an average number of 176 slices, and were collected either from the Coronacases Initiative or the Radiopaedia with a spatial resolution of $512\times512$ for Coronacases Initiative and $630\times630$ for Radiopaedia.

For fair comparisons, we followed the preprocessing strategies introduced in \citep{muller2020automated}. Specifically, based on the Hounsfield units (HU), the raw pixel intensity values within [-1250, 250] are kept, which cover the range of valid lung regions ([-1000, -700]) and Covid-19 infection regions ([50, 100]). The clipping was only applied on the Coronacases Initiative CTs. We then performed z-score normalization on the clipped data to obtain the standardized gray-scale values. To resample the volumes for better neural network training, we adopted a target spacing of $1.58\times1.58\times2.70\ mm^3$ and resampled all data to the median shape of $267\times254\times104\ mm^3$.

\subsubsection{Heart and Hippocampus}
The Heart and Hippocampus datasets are parts of the Medical Segmentation Decathlon (MSD) \citep{simpson2019large}, which is a commonly used 3D medical image segmentation benchmark. The Heart dataset contains 20 training MRI volumes scanned over the entire heart during a single cardiac phase. Images were originally obtained with the voxel resolution $1.25\times1.25\times2.7mm^3$. The target of this task is the segmentation of left atrium, resulting in a binary segmentation task. The Hippocampus dataset includes MRI volumes acquired in 90 healthy adults and 105 adults with a non-affective psychotic disorder (56 schizophrenia,
32 schizoaffective disorder, and 17 schizophreniform disorder). All collected volumes are in the resolution of $1.0\times1.0\times1.0 mm^3$ with labels for anterior and posterior regions, resulting in a three-class segmentation task. 

Our pre-processing strategy follows a common protocol \citep{isensee2018nnu} which includes 3 main steps. First, due to the large number of ineffective signals (i.e., background) exist in the datasets, feeding the raw volumes directly to the networks is computational expensive and time consuming. We thus cropped the data to the region of nonzero values with at least one non-background voxel presented in each volume. Second, similar to the Covid-19 dataset, we resampled voxel spacing to ease network training. For the Heart dataset, raw volumes were resampled to a target shape of $115\times320\times232$ voxels. For the Hippocampus dataset, the target shape is $36\times50\times35$ voxels. Third order spline interpolation was used for input volumes and nearest neighbor interpolation was used for the ground truth masks. Finally, all volumes were z-score normalized individually before being fed into the networks. When the cropping step drops more than 1/4 of the average sizes, we applied the normalization only within the nonzero regions.


\subsection{Evaluation Metrics}

Multiple evaluation metrics are used to quantify the experimental results. For the 2D datasets (MoNuSeg, TNBC, and CHAOS), we rely on mean Intersection over Union (mIoU) and Dice coefficient (DICE) scores to evaluate the models' performances. The two metrics are good indicators of the rate of true positive predictions. For the 3D datasets (Covid-19, Heart, and Hippocampus), metrics including the DICE score, sensitivity and specificity are used. Additionally, total number of network parameters and Multiple-ACcumulate operations (MACs) are also reported for network complexity.


For the datasets without standardized train/test splits (CHAOS, Covid-19, Heart and Hippocampus), we adopted 5-fold cross validation (CV) on the training data. 

\subsection{Statistical Analysis}
For all comparison experiments, we report the average results along with the standard deviation of multiple independent runs for each model. For the datasets with clear testing set split (MoNuSeg and TNBC), we repeat the evaluation three times with different random seeds. For the datasets that require cross-validation (CHAOS, Covid-19, Heart and Hippocampus), we compute the statistics across all validation folds. 

Additionally, we perform two-tail paired t-test to analyze the statistical significance between our methods and the competing methods. Smaller p-values indicate greater statistical significance of improvement that our methods have achieved. 


\begin{figure}[t]
	\includegraphics[width=\linewidth]{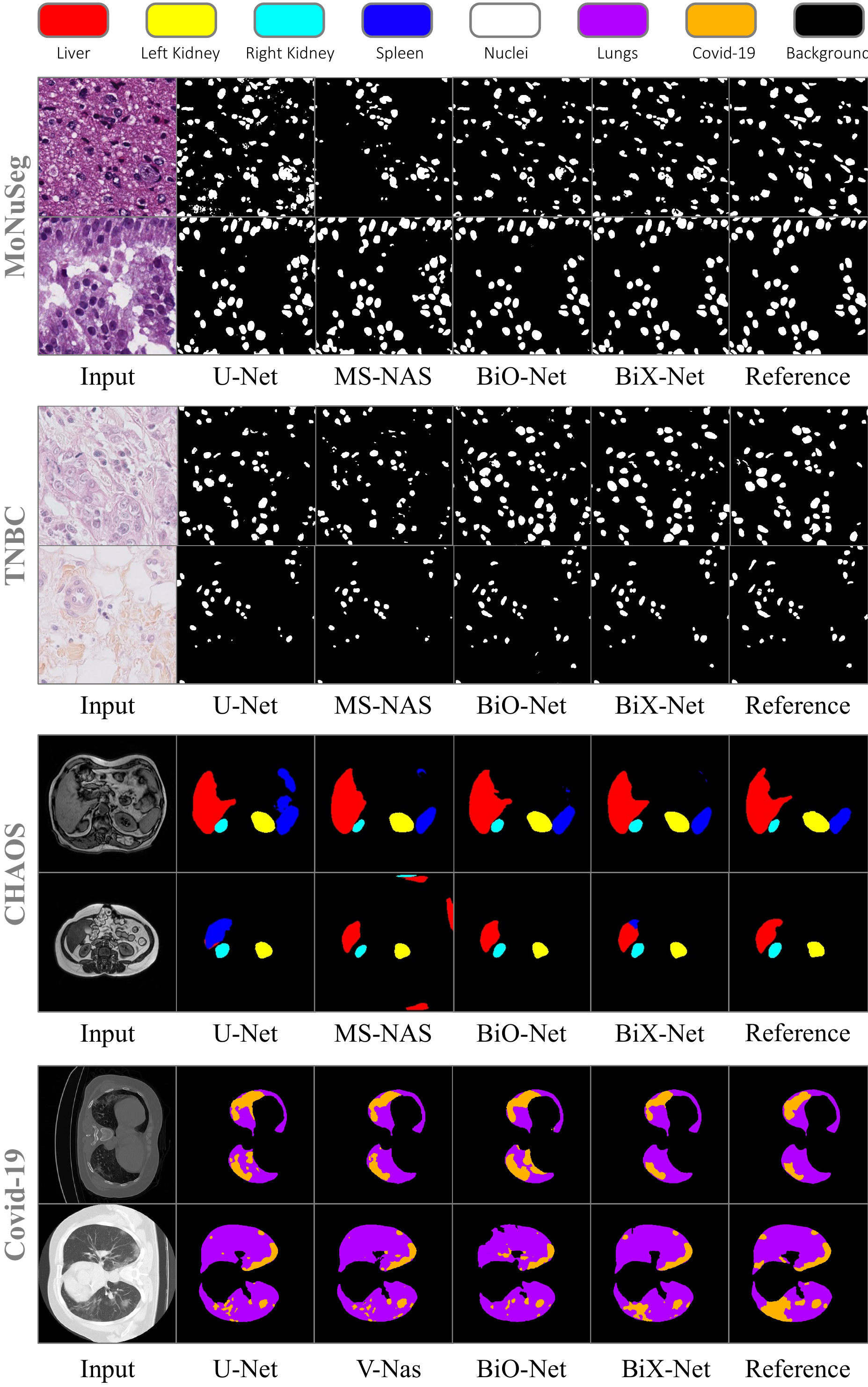}
	\caption{Qualitative comparisons between our BiO-Net, BiX-Net and other competing methods.} \label{qualitative_results}
\end{figure}

\begin{figure*}[t]
	\includegraphics[width=\linewidth]{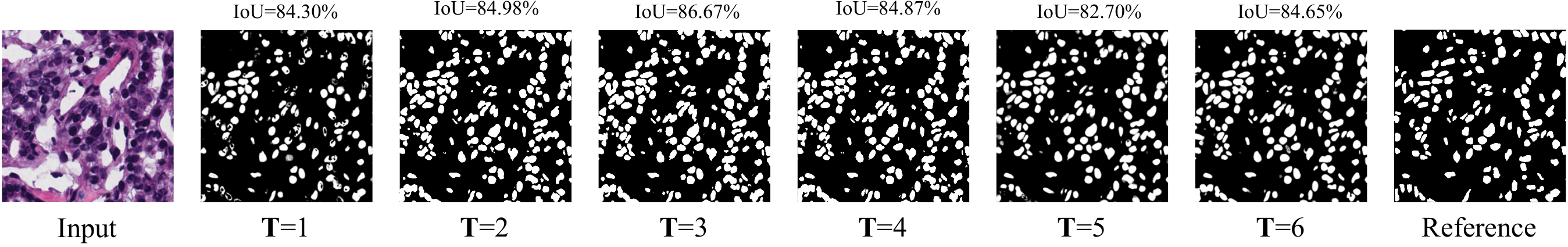}
	\caption{Qualitative results on different recurrence time $\mathbf{T}$ in BiO-Net. Instance IoU scores are shown as well.} \label{rec_results}
\end{figure*}

	\begin{table*}[t]
		\caption{Quantitative comparison on the Covid-19 dataset. For each metric, the best result is in bold. }\label{comp:covid}
		\centering
		\setlength\tabcolsep{1.15em}
		\begin{tabular}{{l|c c c|c c c |c c}}
			\toprule
			\multicolumn{1}{c}{} & \multicolumn{3}{c}{Lungs} & \multicolumn{3}{c}{Infection} & \\
			\hline
			Methods &  DICE (\%) & Sens. (\%) & Spec. (\%) & DICE (\%) & Sens. (\%) & Spec. (\%) & \#Params \\
			\hline
			\hline
			3D U-Net &95.6$\pm$2.6& 95.6$\pm$2.4 & 99.8$\pm$0.2 &76.1$\pm$10.7 &73.0$\pm$15.3 & 99.9$\pm$0.0&22.6 M \\
			V-Nas & 93.3$\pm$4.4 & 94.8$\pm$3.0 & 99.8$\pm$0.0 & 76.0$\pm$11.3 & 73.7$\pm$15.8 & 99.9$\pm$0.0 & 1.6 M\\
			UXNet & \textbf{95.8$\pm$1.4} & 95.3$\pm$1.7 & 99.8$\pm$0.0 & 75.4$\pm$6.1\space\space\space & 72.4$\pm$11.1 & 99.9$\pm$0.0& 9.5 M\\
			\hline
			BiO-Net & 91.0$\pm$7.7&92.4$\pm$8.2 &99.7$\pm$0.2 & 75.1$\pm$10.5& \textbf{77.8$\pm$14.0} &99.8$\pm$0.2& 44.9 M \\
			BiX-Net & 94.8$\pm$1.8& \textbf{96.1$\pm$0.7} & \textbf{99.8$\pm$0.0} & \textbf{78.1$\pm$8.1}\space\space\space& 76.6$\pm$10.2& \textbf{99.9$\pm$0.0} & \textbf{1.4 M}\\
			\bottomrule
		\end{tabular}
	\end{table*}
	
	\begin{table*}[h]
		\caption{Quantitative comparison on the Heart and Hippocampus datasets. For each metric, the best result is in bold. }\label{comp:msd}
		\centering
		\begin{tabular}{{l|c c c|c c c | c c c}}
			\toprule
			 \multicolumn{1}{c}{}& \multicolumn{3}{c}{Heart} & \multicolumn{6}{c}{Hippocampus} \\
			 \hline
			Class & \multicolumn{3}{c|}{1}&\multicolumn{3}{c}{1}&\multicolumn{3}{c}{2}\\
			
			Methods &  DICE (\%) & Sens. (\%) & Spec. (\%) & DICE (\%) & Sens. (\%) & Spec. (\%) & DICE (\%) & Sens. (\%) & Spec. (\%)\\
			\hline
			\hline
			nnU-Net$^{1}$ & 93.3$\pm$0.8 & 93.9$\pm$1.4 & 92.9$\pm$2.0 & 89.8$\pm$0.4 & 89.5$\pm$1.0 & 90.3$\pm$0.7 &88.1$\pm$0.2&\textbf{88.0$\pm$0.4}&88.4$\pm$0.5 \\
			\hline
			BiO-Net & \textbf{93.4$\pm$0.8} & \textbf{94.0$\pm$1.5} &93.0$\pm$2.1 & \textbf{89.9$\pm$0.4}& \textbf{89.8$\pm$0.9} &90.2$\pm$0.8&\textbf{88.2$\pm$0.3}&\textbf{88.0$\pm$0.5}&88.6$\pm$0.6 \\
			BiX-Net & \textbf{93.4$\pm$1.4} & 93.9$\pm$2.1 & \textbf{93.1$\pm$2.2} & \textbf{89.9$\pm$0.2} & 89.5$\pm$0.9& \textbf{90.6$\pm$0.7} & \textbf{88.2$\pm$0.2}&\textbf{88.0$\pm$0.6}&\textbf{88.7$\pm$0.9}\\
			\bottomrule
		\end{tabular}
		\footnotesize
		\begin{tablenotes}
		\setlength{\itemindent}{0.in}
		    \item $^{1}$ nnU-Net \citep{isensee2018nnu} modifies the original U-Net structure and utilizes different network variants for different tasks. Here we use their official model checkpoints for fair comparison.
		\end{tablenotes}
	\end{table*}

\section{Implementation Details} \label{exp}

Our empirical experiments are designed as follows: \textbf{(1)} We firstly compare our BiO-Net and BiX-Net with other state-of-the-art segmentation networks on the 2D nuclei segmentation tasks and the multi-class organ segmentation task. \textbf{(2)} Then, we extend our networks to 3D segmentation tasks, compare to the counterparts that particularly designed for 3D data. \textbf{(3)} Extensive ablation studies are conducted on our BiO-Net and BiX-NAS separately using the nuclei segmentation datasets. Note that we utilized only the MoNuSeg dataset for searching the BiX-Net, and the same architecture is then transferred to all other 2D and 3D tasks.


\subsection{Network Implementation Details}
\paragraph{BiO-Net implementation details} Unless explicitly specified, our BiO-Net is constructed with an encoding depth of $\mathbf{L}=4$ and a backward skip connection built at each stage of the network, except for pre-processing and post-processing blocks. As a balance between computation efficiency and segmentation performance, we set the recurrence time $\mathbf{T}=3$, such that all backward skip connections are triggered 2 times (relevant studies are presented in Sec. \ref{ablation2}). 

\paragraph{BiX-Net searching details} Following the same set up as BiO-Net, we construct the SuperNet BiO-Net++ with $\mathbf{L}=4$ and $\mathbf{T}=3$ as well, resulting in the same hierarchy in the finally searched BiX-Net. In Phase1 search process of our BiX-NAS, we jointly optimize the weights of SuperNet and the selection matrices using the same optimizer, rather than optimized separately \citep{liu2018darts}. The Phase1 searching process took roughly 0.09 GPU-Day. In Phase2, there were in total 5 searching iterations when $\mathbf{T}=3$. At each iteration, for each retained architecture from the preceding iteration, we sampled $\mathbf{s}=15$ different skip sets to form the new $\mathbf{P}$. Due to GPU memory limitation, we applied a channel expansion parameter of $\mathbf{W}=0.75$ and only retained less than three final architectures ranked by IoU at the Pareto front per search iteration. Our Phase2 searching process consumed 0.37 GPU-Day. After Phase2 searching finished, we inserted the searched bi-directional skip connections into a re-initialized encoder-decoder architecture. For the blocks that have been discarded by our search algorithm, we abandoned all computations (i.e., convolutions, normalizations and activations) in such blocks but still resized the incoming features to the corresponding scale.

\subsection{Training Details}
All our training configurations followed common usages without any explicit tuning. The competing methods were trained under the identical settings for fair comparison. \textcolor{black}{Note that it is a common protocol in the NAS community that searching is performed on a small dataset only and the searched network is then generalized to larger domains \citep{bender2018understanding, liu2018darts, yan2020ms}. Such protocol not only demands much lower search costs but also serves as good evaluation for the generalization ability of the proposed NAS algorithms. Therefore, we strictly followed such protocol and searched our BiX-Net on the small-scale MoNuSeg dataset only and transferred the same architecture to more complex 2D and 3D tasks.}


\subsubsection{MoNuSeg and TNBC}

Standard data augmentation strategies were applied to the input data on-the-fly including: random rotation (within the range [-15$^\circ$, +15$^\circ$]), random translation (in both x- and y-directions; within the range of [-5\%, 5\%]), random shearing, random zooming (within the range [0, 0.2]), and random flipping (both horizontally and vertically). We train all models using the Adam \citep{kingma:adam} optimizer for 300 epochs with batch size 2. For manually crafted networks including BiO-Net, the initial learning rate was set to 0.01 with a decay rate of 0.003 at every epoch. For all NAS searched networks including BiX-Net, the initial learning rate was set to 0.001 with the same decay rate. 

During the Phase1 search, we trained the BiO-Net++ SuperNet for 300 epochs with a base learning rate of 0.001 and a decay rate of 3e-3. During the Phase2 search, at each searching iteration, we trained the sampled candidate networks 40 epochs starting from a learning rate of 0.001 and then decayed by 0.1 every 10 epochs. 

\subsubsection{CHAOS}

Since there are many more training slices in the CHAOS dataset than the nuclei segmentation datasets, we lowered the initial learning rate to 0.001, which remains consistent across both handcrafted and NAS searched networks. The learning rate was cosine-annealing scheduled to 10$^{-7}$ in 300 epochs. Data augmentations including random spatial translation within the range of [-30\%, 30\%], random flipping, and random elastic transformation with 1.5 alpha and 0.07 sigma were used. We trained all models using the SGD optimizer for 300 epochs with momentum 0.9, weight decay 0.0001 and batch size 16. 

\subsubsection{Covid-19}

Differing from the above 2D segmentation tasks, we employed the Adam optimizer to minimize the both the cross entropy loss and the Tversky loss \citep{salehi2017tversky} with 0.5 alpha, 0.5 beta and batch size 2. The initial learning rate is set to 0.0003 for BiO-Net and BiX-Net and 0.001 for other competing methods (recommended settings in \citep{muller2020automated}). Learning rates were scaled down by 0.1 once learning stagnates with a patience of 20 epochs. The minimum limit of the learning rate is 1e$^{-5}$. Following \citep{muller2020automated}, the data augmentation process includes: random mirroring, random elastic transformations, random rotations, random brightness enhancement, random contrast, random gamma corrections, and random Gaussian noise injections. The probability of triggering each of the augmentation strategy is 15\%. 

Due to GPU memory limitations, the networks were trained with $160\times160\times80$ voxel patches randomly sampled from the raw volumes. During inference, we sampled the patches via the sliding window strategy enabling an overlap of half the patch size ($80\times80\times40$ voxels) to stabilize boundary predictions. 

\subsubsection{Heart and Hippocampus}

For fair comparisons, we followed the same training settings and augmentation strategies introduced in \citep{isensee2018nnu}. Adam optimizer with an initial learning rate of 0.0003 was used across all experiments. We trained the networks with randomly sampled patches from the pre-cropped volumes. For better stabilizing optimization, we ensure that at least 1/3 of each batch contains data in foreground class. We denote 250 training steps as one epoch, and set the maximum epoch limit to be 1000. \textcolor{black}{Similar to the Covid-19 dataset}, the learning rate was reduced by a factor of 5 when the training loss does not change by at least 0.005. \textcolor{black}{We stopped the training when the learning rate fell below 0.000001 and no significant change was observed in the training loss.} The augmentation strategies include random rotations, random scaling, random elastic deformations, gamma corrections, and mirroring.

\section{Results} \label{results}
In this section, we present the experimental results obtained following the implementation details introduced in Sec. \ref{exp}.

\subsection{Comparison Results}

\subsubsection{MoNuSeg and TNBC}


The results on the MoNuSeg and TNBC datasets are reported in Table \ref{comp:nuclei}. We compare our networks with three types of counterparts: \textbf{(1)} state-of-the-art U-Net variants: U-Net++ \citep{zhou2018unetpp}, Attention U-Net \citep{oktay2018attention}; \textbf{(2)} recurrent networks: R-UNet \citep{wang2019recurrent}, R2U-Net \citep{wang2019recurrent}; \textbf{(3)} NAS searched networks: NAS-UNet \citep{weng2019unet}, AutoDeepLab \citep{liu2019auto}, MS-NAS \citep{yan2020ms}.

Although the U-Net variants demonstrate good performances on the validation set of MoNuSeg, they require a large number of network parameters. When validated on the TNBC dataset that has different data distributions, the generalization ability of such networks becomes limited. One advantage of recurrent networks is that the number of parameters remains the same with increased recurrence time. By recursing through the same network layers, performance is improved on both MoNuSeg and TNBC datasets. However, as discussed earlier, the computational costs (MACs) inevitably escalate with more recurrence time. With a good balance between segmentation performance and network complexity, the NAS searched networks yield better generalization results on the TNBC dataset than the manually crafted U-Net variants. 

By incorporating our proposed bi-directional skip connections, BiO-Net demonstrates superior performances on both datasets. However, when $\mathbf{T}=3$, an increase of 75.7\% MACs is observed against the plain U-Net baseline. As expected, our proposed BiX-NAS successfully discovered an optimal network instance with minimum computational costs. Our resource-aware BiX-Net achieves on par performance to the heavy BiO-Net on the MoNuSeg validation set and even better generalization results on the TNBC dataset.

\subsubsection{CHAOS}

Giving the great results achieved on binary segmentation tasks, we then examined if the same performance gain can be obtained on a multi-class segmentation task. The results on organ segmentation tasks for the CHAOS dataset are shown in Table \ref{comp:chaos}. Compared to other competing methods, our BiO-Net and BiX-Net demonstrate superior performances on all metrics across all classes. \textcolor{black}{Noteworthy, BiX-Net is able to achieve the best results on a binary nuclei segmentation dataset when searched on another dataset in the same domain. However, the searched BiX-Net appears to be less effective when transferred to the CHAOS dataset for multi-organ segmentation. Such performance gap indicates a common generalization bottleneck between different domains. With $4\times$ lower computational complexity and $34.9\times$ fewer parameters than BiO-Net, BiX-Net still achieved the second best results that surpass all other comparison methods.}


\begin{table*} [t]
		\caption{Ablative results on backward skip connections in BiO-Nets. IoU (DICE) and number of parameters are reported. \textcolor{black}{Our adopted BiO-Net set up is presented as \textbf{reference} with the results stated in bold.}}\label{tab:ablation1}
		\setlength\tabcolsep{1.1em}
		\centering
		\begin{tabular}{{c| c c c |c c c | c}}
				\toprule
				\multicolumn{1}{c}{} & \multicolumn{3}{c}{MoNuSeg (\%)} & \multicolumn{3}{c}{TNBC (\%)} & \\
				\hline
				&  $\mathbf{T}=1$ & $\mathbf{T}=2$ & $\mathbf{T}=3$ &  $\mathbf{T}=1$ & $\mathbf{T}=2$ & $\mathbf{T}=3$ & \#Params \\
				\hline
				\hline
				\textbf{Reference$^1$} & \textbf{68.0(80.3)} & \textbf{69.4(81.6)} & \textbf{69.9(82.0)} & \textbf{45.6(60.8)} & \textbf{54.8(69.3)} & \textbf{61.8(75.1)} & 15.0 M\\
				\hline				
				$\mathbf{N}=1.25$ & 68.5(81.3) & 69.8(81.9)& 69.5(81.7)& 49.0(63.7)& 55.7(69.7)&62.3(75.8)&23.5 M\\
				$\mathbf{N}=0.75$& 67.6(80.0)& 67.8(80.5)& 69.1(81.5)&51.6(66.1)&57.1(71.0) &59.8(73.8)& 8.5 M\\
				$\mathbf{N}=0.50$ & 66.8(79.2)& 68.0(80.6) & 69.1(81.4)& 49.1(64.4) & 54.3(67.9)&61.1(74.2) & 3.8 M\\
				$\mathbf{M}=0.25$ & 66.7(79.1)& 67.8(80.4)& 67.7(80.4)& 52.4(67.4)&53.5(67.8) &57.5(71.0) & 0.9 M   \\
				\hline
				$\mathbf{W}=3$ & 68.0(80.3) & 69.4(81.7) & 68.8(81.4) &45.6(60.8) &51.0(65.6)& 62.0(75.7) &15.0 M \\
				$\mathbf{W}=2$& 68.0(80.3) & 67.2(80.1)& 68.6(81.3) &  45.6(60.8) &52.7(67.9)& 60.1(74.2) & 14.9 M \\
				\hline
				$\mathbf{L}=3$& 68.1(80.6) & 67.9(80.5)&68.9(81.2) & 61.3(74.2) & 59.4(73.3)& 61.5(74.1) &3.8 M \\
				$\mathbf{L}=2$& 69.0(81.0) & 69.5(81.7) & 69.7(81.8)& 59.6(73.4)&64.7(77.5)&59.6(73.5)& 0.9 M\\
				\bottomrule
		\end{tabular}
		\footnotesize
		\begin{tablenotes}
		\setlength{\itemindent}{0.in}
		    \item $^{1}$ Our adopted BiO-Net set up with $\mathbf{N}=1.0, \mathbf{W}=4$, and $\mathbf{L}=4$.
		\end{tablenotes}
\end{table*}
	
\begin{figure*}[t]
	\includegraphics[width=\textwidth]{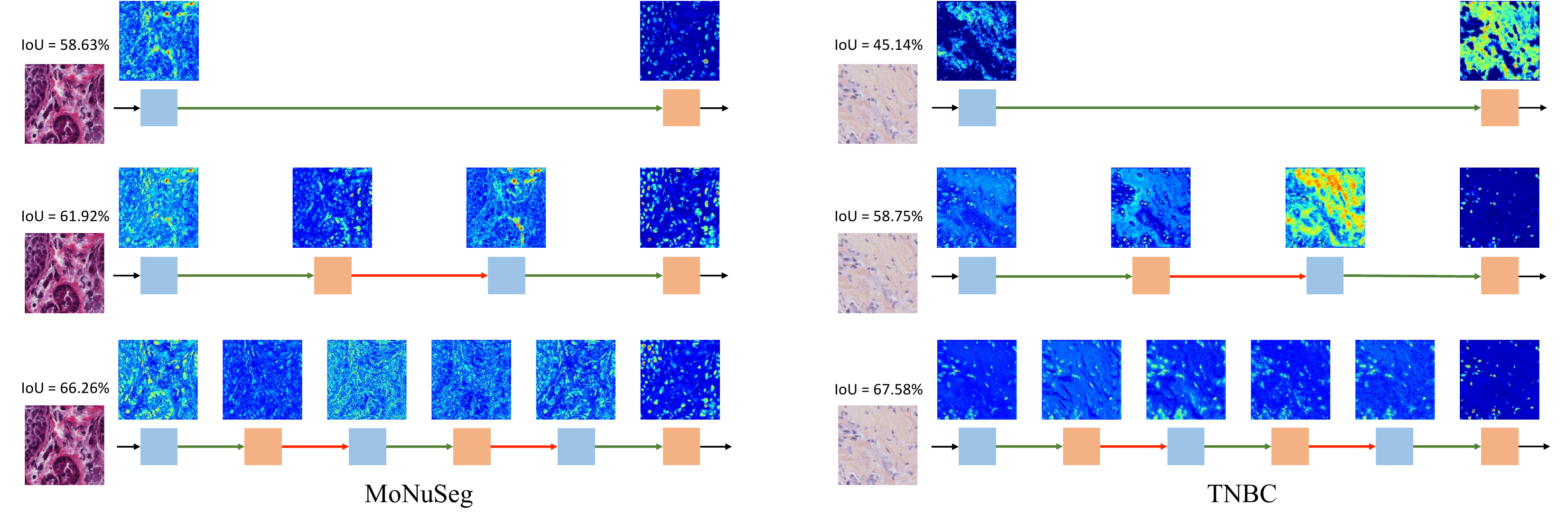}
	\caption{Visualizations of the averaged feature maps obtained at different iterations. Instance IoU scores for the input image are reported.} \label{feature_map}
\end{figure*}

\subsubsection{Covid-19}

After the evaluations on 2D datasets, we then validate our BiO-Net and BiX-Net on 3D segmentation tasks. First, we present the quantitative results on the Covid-19 dataset in Table \ref{comp:covid}. The compared methods include: the baseline network 3D U-Net \citep{cciccek20163d}, the state-of-the-art NAS searched networks V-Nas \citep{zhu2019v} and UXNet \citep{ji2020uxnet}. Since 3D segmentation is much more challenging than the 2D tasks, the results achieved by our BiO-Net and BiX-Net are not statistically significant (i.e., p-values $>$ 0.05) compared to other methods and therefore no cells are highlighted in Table \ref{comp:covid}. However, except for the DICE score on lung segmentations, our proposed networks yield the highest results on all other metrics. Our findings are as follows: \textbf{(1)} A naive bi-directional network recursion for 3D segmentation cannot bring the same performance enhancement as in 2D tasks. This can be observed by comparing BiO-Net with the non-recursable 3D U-Net. Our understanding is that 3D features are more complex than the 2D ones. Therefore, the network would easily saturate with the same capacity. The same behaviour can also be observed in Sec. \ref{ablation1} where BiO-Net can hardly benefit from bi-directional skip connections with limited parameters. \textbf{(2)} BiX-Net stands out with superior results on almost all metrics. Noticeably, BiX-Net was searched on MoNuSeg only, and the great performances prove the generalization ability to transfer the identical architecture to tasks in significant different domains.

\begin{figure*}[t]
	\includegraphics[width=\linewidth]{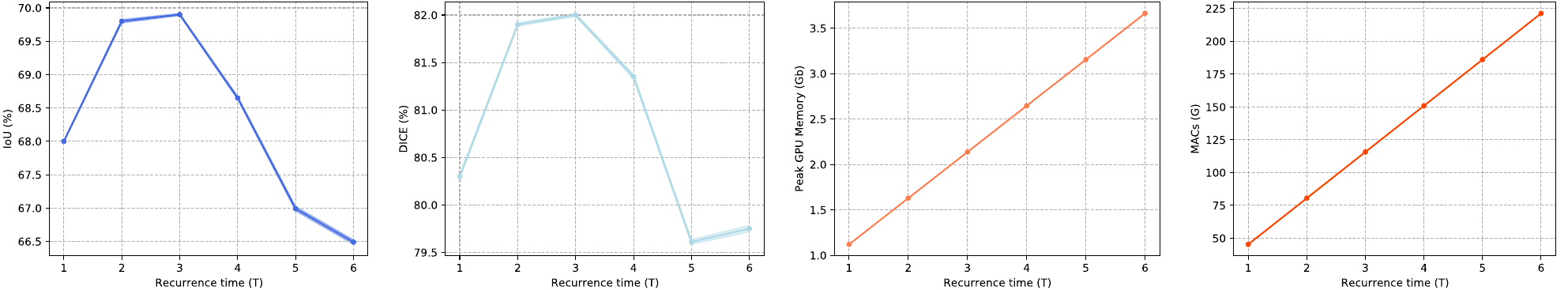}
	\caption{Ablative results on different recurrence time $\mathbf{T}$ in BiO-Net. GPU memory is measured during network inference.} \label{ablation_rec}
\end{figure*}

	\begin{table*}[t]
		\caption{Ablative results on different searching phases. \textcolor{black}{Best results in each metric are in bold.}}\label{tab:ablation4}
		\centering
		\setlength\tabcolsep{0.7em}
		\begin{tabular}{l|c c|c c|c r | c r}
			\toprule
			\multicolumn{1}{c}{} & \multicolumn{2}{c}{MoNuSeg} & \multicolumn{2}{c}{TNBC}& 	\multicolumn{3}{c}{}\\
			\hline
			Methods &  IoU (\%) & DICE (\%) & IoU (\%) & DICE (\%) &\#Params& Overhead$^1$ & MACs & Overhead$^1$\\
			\hline
			\hline
			BiO-Net &  69.9$\pm$0.2 &  82.0$\pm$0.2 &  62.2$\pm$0.4 & 75.8$\pm$0.5 & 14.99 M&3845\% & 115.67 G & 313\%\\
			BiO-Net++ &\textbf{70.0$\pm$0.3}&\textbf{82.2$\pm$0.3}&67.5$\pm$0.4&80.4$\pm$0.5&0.43 M& 13\%&34.36 G & 23\%\\
			Phase1 searched &69.8$\pm$0.2&82.1$\pm$0.2&66.8$\pm$0.6&80.1$\pm$0.4&0.43 M& 13\%&31.41 G & 12\%\\
			BiX-Net & 69.9$\pm$0.3 & \textbf{82.2$\pm$0.2} & \textbf{68.0$\pm$0.4} & \textbf{80.8$\pm$0.3} & \textbf{0.38 M}&0\%& \textbf{28.00 G}&0\%\\
			\bottomrule
		\end{tabular}
		\footnotesize
		\begin{tablenotes}
         \item $^{1}$ Overhead compared to BiX-Net.
    \end{tablenotes}
	\end{table*}

\subsubsection{Results on Heart and Hippocampus}

Finally, our proposed methods were evaluated on the Heart and Hippocampus datasets from the medical segmentation decathlon. We compare our networks with the second place in the MSD official leader board: nnUNet, and used their provided model checkpoints to produce their predictions. The comparisons are reported in Table \ref{comp:msd}. Due to the difficulty of the tasks, our methods only demonstrate marginal improvements. However, such improvements can be observed on nearly all the metrics. Noteworthy, as a computational efficient mobile network, our BiX-Net is able to achieve the state-of-the-art results.

\subsubsection{Qualitative Results}

For more intuitive comparisons, we present two qualitative segmentation results on each of the MoNuSeg, TNBC, CHAOS, and Covid-19 datasets in Fig.  \ref{qualitative_results}. Clearly, our BiO-Net and BiX-Net produce the most accurate segmentation masks. According to the visualization results, there are two primary advantages of our methods: \textbf{(1)} Our methods include less noise in the final prediction (MoNuSeg first case, TNBC first case). \textbf{(2)} Our methods produce the least false positive predictions (CHAOS).


\subsection{Ablation Studies}

We conducted extensive ablative experiments to inspect the behaviours of our methods under different setups. Unless explicitly specified, we used the nuclei segmentation datasets for the ablation studies.

	

\subsubsection{Effectiveness of Bi-directional Skip Connections} \label{ablation1}
To better investigate the effectiveness of our backward skip connections, we here study the impact of 3 different factors under different scenarios: different network size controlled by a channel expansion parameter $\mathbf{M}$, different number of backward skip connections used from the deepest encoding level $\mathbf{W}$ and different encoding depth of the network $\mathbf{L}$. For more comprehensive studies, we report the ablative results on different recurrence time from $\mathbf{T}=1$ to $\mathbf{T}=3$. We ignore all other influential factors and use the plain BiO-Net for evaluation. 

We present the ablation results impacted by different factors in Table \ref{tab:ablation1}. With the increasing recurrence time, performance improvements can be easily observed on nearly all experiments. Specifically, we observe that: \textbf{(1)} When the number of training parameters is restricted (e.g., $\mathbf{M}$=0.25), our BiO-Net could hardly benefit from reusing convolutional layers through bi-directional skip connections. This observation aligns with the ones spotted in 3D segmentation tasks. \textbf{(2)} Without significant reductions on training parameters, using less backward skip connection while maintaining the same forward ones (e.g., $\mathbf{W}$=3) causes slight performance drop on MoNuSeg when $\mathbf{T}$ increases. However, the generalization results on TNBC appear even better than the reference setting, with further 0.6 improvement on the DICE score. \textbf{(3)} Surprisingly, building bi-directional skips on a shallower network (e.g., $\mathbf{L}$=3) not only reduces the total parameters but leads to impressive results on the MoNuSeg datasets. The same results performance boost was however not observed on the TNBC dataset. 

Given the quantitative improvements brought by our bi-directional skip connections, one may wonder if the additional recurrence could actually generate meaningful features, instead of generating replicated or uninformative values. We visualize the averaged feature maps inferred at different recurrence steps in Fig. \ref{feature_map}. The averaged feature maps vary along different time steps, validating that the recurrence inference through bi-directional skip connections is indeed able to generate indicative feature representations.

\subsubsection{Impact of Recurrence Time}\label{ablation2}

With the help of backward skip connections, our network is able to recurse through the encoder and decoder for an iterative refinement of the features. Under the memory constraints, we are interested in inspecting how recurrence time $\mathbf{T}$ impacts the final segmentation results. To this end, we conducted experiments on varying the recurrence time 
for $\mathbf{T}\in[1,6]$. In addition to the metrics measuring segmentation performances, extra metrics including peak inference GPU memory usage and MACs are reported to demonstrate the computational burdens change along with the increasing recurrent time. 

Although recursing through encoder and decoder earns extra performance improvement, we hypothesize that our BiO-Net would eventually saturate giving the fixed training parameters, and results in performance drops after a certain time of recurrence. As shown in Fig. \ref{ablation_rec}, segmentation performances of BiO-Net consistently increase when $\mathbf{T}\leq3$ and reach the greatest results at $\mathbf{T}=3$. However, a longer network recurrence harms the feature representation and leads to even worse results. When $\mathbf{T}\geq5$ both IoU and DICE scores are even poorer than $\mathbf{T}=1$. Also, we note the peak GPU memory usage and MACs are linearly increased along with the recurrence time, hence, unrestricted recurrent inference is impractical. In Fig. \ref{rec_results}, we compare the network predictions under different maximum $\mathbf{T}$. The above observations validate the demand of a resource-aware and efficient substitute, which motivate our BiX-Net design. As a good trade-off between network performance and computation costs, we adopt $\mathbf{T}=3$ to be our default setting and search BiX-Net across 6 extraction stages only (3 encoders + 3 decoders).
	
	\begin{table*}[t]
		\caption{Ablative results on different Phase2 search strategies. \textcolor{black}{Best results in each metric are in bold.}}\label{tab:ablation5}
		\centering
		\setlength\tabcolsep{0.95em}
		\begin{tabular}{{l|c c|c c|c | c}}
			\toprule
			\multicolumn{1}{c}{} & \multicolumn{2}{c}{MoNuSeg} & \multicolumn{2}{c}{TNBC}& 	\multicolumn{2}{c}{}\\
			\hline
			Phase2 search strategies &  IoU (\%) & DICE (\%) & IoU (\%) & DICE (\%) &Search time& MACs\\
			\hline
			\hline
			Random search & 67.5$\pm$0.4 &80.4$\pm$0.4  & 57.4$\pm$0.5 & 69.8$\pm$0.6 & 0.69 GPU day & 30.40 G\\
			Independent training & 68.4$\pm$0.3 &81.1$\pm$0.4  & 60.9$\pm$0.6 & 73.6$\pm$0.7 & 0.73 GPU day& 29.68 G\\
			Progressive evolutionary search & \textbf{69.9$\pm$0.3} & \textbf{82.2$\pm$0.2}  & \textbf{68.0$\pm$0.4} & \textbf{80.8$\pm$0.3} & \textbf{0.37 GPU day}&\textbf{28.00 G}\\
			\bottomrule
		\end{tabular}
	\end{table*}

\subsubsection{Necessity of the Progressive Evolutionary Search}\label{ablation4}
We claimed in Sec. \ref{exnas} that the Phase1 searched architecture still has computation redundancies and a following evolution-based Phase2 search can spot more effective and efficient sub-set skips. We conducted two extensive experiments to prove the necessity of the proposed progressive evolutionary search by training and evaluating the SuperNet BiO-Net++ and the Phase1 searched architecture directly. 

We report the quantitative results on different intermediate networks in Table \ref{tab:ablation4}. Compared to BiO-Net that is constructed of identical building blocks and fusion functions to the plain U-Net, the modified BiO-Net++ requires only 1/34.86 parameters and 1/4 computations. With the help of multi-scale feature fusions, BiO-Net++ achieves even better results on both datasets. After performing BiX-NAS to reduce the skip redundancies in BiO-Net++, our Phase1 searched architecture reduces the computations by 8.6\% further with nearly no influences on the segmentation results. Finally, our Phase2 searched BiX-Net shrinks the computation of BiO-Net++ by 17\% and still achieves on par results.
	
\subsubsection{Efficiency of the Progressive Evolutionary Search}
\label{ablation5}

We theoretically analyzed that our proposed searching method is computational efficient and could potentially lead to better results by adhering the skip fairness concept. Here we provide empirical validations on our claims: First, we compare two counterpart strategies: \textbf{(1)} randomly search skip connections across all extraction stages from the SuperNet (i.e., without progressive evolution); and \textbf{(2)} progressively search skip connections and train each network instance independently (i.e., without adhering skip fairness). Then, we construct the optimal network instances searched by the above strategies, and compare the searching cost and retraining performances on the MoNuSeg and TNBC datasets. 


As shown in Table \ref{tab:ablation5}, without an evolution schema, searching directly on randomly sampled skip connections leads to a sub-optimal architecture instance. The results on both datasets are the lowest and the search process requires a great amount of time. As discussed earlier, a limited number of samples will not suffice to completely explore the vast search space and the true optimal instance is unlikely to be covered. 

By evolving the skip connections from the last pair of extraction stages to the first, the final network instance achieves much better performances on both datasets. However, independent instance training violates the skip fairness and consumes an unneglectable search time, thus resulting in inferior segmentation performances and exceeding search time.

Eventually, with the proposed progressive evolutionary search, our BiX-Net achieves the best performances in terms of all metrics. By sharing the head network among the candidates, the skip fairness is perfectly met and greatly reduces the overall search time. Compared to other search strategies, our proposed algorithm only demands about half of the time, proving the efficiency and effectiveness of our progressive evolutionary search.

\section{Conclusion and Future Work}
In this work, we studied the skip schema in encoder-decoder architectures. Paired to forward skip connections, we proposed backward skip connections to ship decoded features back to encoder. The bi-directional skip connections are incorporated into a simple encoder-decoder architecture, namely BiO-Net that reuses network parameters in a recurrent manner. We analyzed the complexity overhead of the plain BiO-Net and designed the economic multi-scale BiO-Net++ with dense skip connections. A two-phase NAS algorithm, BiX-NAS, was further proposed to automatically search for the optimal sub-set skip connections in BiO-Net++. At the first phase, we utilized a novel selection matrix to narrow down the search space. Our modification supports a flexible output of skips and unifies the forward behaviours during both searching and network inference. At the second phase, we progressively searched for the optimal skip connections between every pair of encoder and decoder. The finally searched BiX-Net yields only 1/39.4 parameters and 1/4.1 computation compared to the plain BiO-Net, but achieves even better segmentation performances on different benchmarks. We evaluated our methods on 3 2D datasets and 3 3D datasets that achieved state-of-the-art performances. Extensive ablation studies were conducted to provide more comprehensive analysis of each of the proposed methods and components.

\textcolor{black}{Toward a natural extension in the future, one can easily upgrade our BiO-Net/BiX-Net to an universal solution by adopting the `domain adaptor' \citep{huang20193d} seamlessly into the networks. Moreover, it is also possible to search for more advanced structures of such `domain adaptor' in every building block by modifying our two-phase BiX-NAS.}

\bibliographystyle{model2-names.bst}\biboptions{authoryear}
\bibliography{refs}



\end{document}